\documentclass{INTERSPEECH2023}

\usepackage[table]{xcolor}
\usepackage[T1]{fontenc}
\usepackage{booktabs}
\usepackage{graphicx}
\usepackage{caption}
\usepackage{subcaption}

\usepackage{bbding}

\definecolor{Gray}{gray}{0.9}

\interspeechcameraready

\title{CommonAccent: Exploring Large Acoustic Pretrained Models for Accent Classification Based on Common Voice}

\name{Juan Zuluaga-Gomez$^{\star,\dagger,\ddagger}$, Sara Ahmed$^{1}$, Danielius Visockas$^{\mathsection}$, Cem Subakan$^{\natural, \sharp, \flat}$.
\thanks{
$^{\star}$Corresponding author. \\
This work was partially supported by \textit{ROXANNE}, a European Union’s Horizon 2020 research project leveraging real-time network, text, and speaker analytics for combating organized crime (grant 833635).
}
}
\address{
  $^{\dagger}$ Idiap Research Institute, Switzerland $^{\ddagger}$ Ecole Polytechnique Federale de Lausanne, Switzerland \\
  $^{1}$ Sketch Recognition Lab, Texas A\&M University, USA \\
  $^{\mathsection}$ Vilnius Gediminas Technical University, Vilnius, Lithuania \\
  $^{\natural}$Université Laval, Canada $^{\sharp}$Concordia University, Canada $^{\flat}$Mila-Québec AI Institute, Canada
}
\email{juan-pablo.zuluaga@idiap.ch}

\begin{document}

\ninept
\maketitle

\begin{abstract}
Despite the recent advancements in Automatic Speech Recognition (ASR), the recognition of accented speech still remains a dominant problem. In order to create more inclusive ASR systems, research has shown that the integration of accent information, as part of a larger ASR framework, can lead to the mitigation of accented speech errors. We address multilingual accent classification through the ECAPA-TDNN and Wav2Vec 2.0/XLSR architectures which have been proven to perform well on a variety of speech-related downstream tasks. We introduce a simple-to-follow recipe aligned to the SpeechBrain toolkit for accent classification based on Common Voice 7.0 (English) and Common Voice 11.0 (Italian, German, and Spanish). Furthermore, we establish new state-of-the-art for English accent classification with as high as 95\% accuracy. We also study the internal categorization of the Wav2Vev 2.0 embeddings through t-SNE, noting that there is a level of clustering based on phonological similarity.\footnote{Our recipe is open-source on the SpeechBrain toolkit, see: \url{https://github.com/speechbrain/speechbrain/tree/develop/recipes}}
\end{abstract}


\section{Introduction}
\label{sec:introduction}

Large acoustic models (LAMs) have become the standard choice for a large variety of downstream tasks, such as automatic speech recognition (ASR) or language identification.\footnote{See more downstream applications in the SUPERB~\cite{superb} website: \url{https://superbbenchmark.org/}} These LAMs are trained using self-supervised learning (SSL) techniques on vast amount of data, often exceeding 50k hours of audio. Their powerful capabilities are evident in the application of datasets such as Librispeech~\cite{panayotov2015librispeech}. However, a significant limitation of these models is that they primarily utilize data spoken in native English or a single accent, thereby neglecting the diversity of accents among speakers. Examples of these pretrained LAMs in only-English are Wav2Vec 2.0 (w2v2)~\cite{wav2vec20} or in a multilingual setup, the w2v2-XLSR model~\cite{xlsr}.

Previous studies have demonstrated that end-to-end ASR models based on pretrained LAMs (e.g. w2v2) exhibit significant performance disparities when applied to non-native English speech. For example, up to $\sim$50\% relative increase in word error rate (WER) has been observed when comparing native (US) and non-native (Malaysian) English speech~\cite{cambara2021english}. As a result, it has become crucial to develop and implement accent-aware or accent-invariant ASR systems. However, only a limited number of studies have specifically addressed this issue. 

In this paper, we study the accent classification problem, which is a critical building block towards accent-aware ASR. Our aim is to provide insight and guidance for evaluating fine-tuned LAMs in a more inclusive manner, highlighting the importance of considering accent variability not only in ASR, but also in different downstream tasks where accent disparity might degrade performances (e.g., spoken language understanding). Specifically, in this work, we introduce a simple-to-follow recipe on the SpeechBrain~\cite{speechbrain} toolkit to perform accent classification based on speech recordings. The recipe fine-tunes either ECAPA-TDNN~\cite{ecapa_tdnn} or w2v2~\cite{wav2vec20} models (also XLSR) in the accent classification task. Our system follows closely the CommonLanguage recipe\footnote{\url{https://github.com/speechbrain/speechbrain/tree/develop/recipes/CommonLanguage}.} available in SpeechBrain. Additionally, we open-source fine-tuned models in the HuggingFace Hub~\cite{huggingface,huggingface_datasets}.

Our recipe utilizes data from Common Voice~\cite{common_voice} dataset in four different languages, namely: English, German, Spanish, and Italian. Our contributions are four-fold as follows:

\begin{itemize}
    \item We open-source \mbox{ECAPA-TDNN}~\cite{ecapa_tdnn} and w2v2~\cite{wav2vec20} fine-tuned models that recognize 16 different accents in the English language, which to the author's knowledge is the largest open-source accent classification system to date. In addition, we also cover 4 accents in German, 6 in Spanish, and 5 in Italian.
    \item We set the first baseline for accent classification based on Common Voice dataset~\cite{common_voice} which, to the author's knowledge, is the largest open-source and free-access acoustic database that provides accent annotations.
    \item We introduce CommonAccent, a subset of Common Voice compiled as a benchmark dataset optimized for accent classification in multiple languages, e.g., English, German, Spanish, and Italian.
    \item Finally, we open-source a recipe, named CommonAccent, in the SpeechBrain tookit~\cite{speechbrain} for performing accent classification based on speech recordings. CommonAccent can be easily adapted to other languages.
\end{itemize}

In Section~\ref{sec:related-work} we cover early work on accent classification. Section~\ref{sec:common_accent} formalizes the data preparation phase for CommonAccent, while also describing the datasets partition. We discuss the main results in Section~\ref{sec:experiments} and~\ref{sec:results} and conclude the paper in Section~\ref{sec:conclusion}.

\section{Related Work}
\label{sec:related-work}

Accents are considered one of the main sources of speech variability. Differences between accents are reflected primarily in three aspects: stress, tone, and length~\cite{shi2021accented}. Accent classification is similar to language identification~\cite{rangan2020exploiting,safitri2016spoken} and speaker verification~\cite{okabe2018attentive,xie2019utterance,nagrani2020voxceleb} as it classifies sequences of speech at the full-length utterance level. Prior research has extensively explored ways to improve the classification of accented speech and its effect on ASR. Early works explored contextual Hidden Markov Model (HMM)-based units~\cite{teixeira1996accent} and the use of formant frequency features into GMM models~\cite{deshpande2005accent}. More recent research has leveraged techniques from a variety of speech technology domains which has lead to promising results. In a recent Interspeech (2020) competition~\cite{shi2021accented},\footnote{Interspeech 2020 challenge: Accented English Speech Recognition Challenge.} the highest performing model used a TDNN based classification network with phonetic posteriorgram (PPG) features as input and TTS (text-to-speech) to augment the training data~\cite{shi2021accented, huang2021aispeech}. Another proposed accent classification network, mined elements from a deep speaker identification framework to make it applicable for accent classification. In detail, they implemented a Convolutional Recurrent Neural Network as a front-end encoder, integrated local features using a Recurrent Neural Network, included a Connectionist Temporal Classification (CTC) based speech recognition auxiliary task, and introduced some strong discriminative loss functions~\cite{wang2020deep}. This work expands upon this area of research by studying more recent deep neural networks, e.g., XLSR, on accent classification for different languages.

\begin{figure}[t]
  \centering
  \includegraphics[width=0.7\columnwidth]{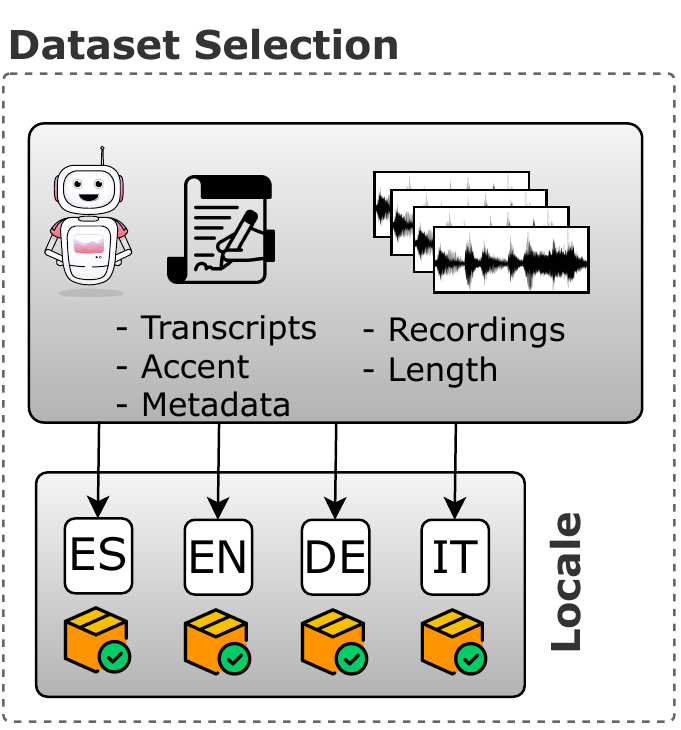}
  \caption{Audio data selection and packaging for English, German, Spanish, and Italian locales of Common Voice dataset.} 
  \label{fig:dataset-selection}
\end{figure}

Accent information, when integrated into an ASR framework, yields promising results. Multitask learning provides a way to transition from maintaining separate acoustic models for each accent to sharing all acoustic model parameters except for accent-specific top-layers~\cite{huang2014multi, chen2015improving}. The inclusion of the speaker's native language data in multitask models has also led to notable decreases in the error rate as well~\cite{Ghorbani2018,Jain2018}. Other areas such as domain expansion~\cite{ghorbani2019domain} and pronunciation modification have also been explored~\cite{radzikowski2021accent}. A CTC model initialized by a w2v2 encoder with LAS rescoring achieved the highest results when compared to other models at the Interspeech accent recognition challenge~\cite{shi2021accented}. In addition to accent, end-to-end systems that account for dialect variations in grammar and vocabulary have also led to error reduction~\cite{imaizumi2020dialect}. 

There are still many challenges facing accented speech and accented ASR. For example, the lack of a standard benchmark, dialectal difference encompassing grammar, vocabulary, and spelling, generalizing to a larger selection of accents, and whether accent-tuning approaches in English are applicable to other languages as well, particularly those that show more dialectal variance~\cite{hinsvark2021accented}. We address these problems in regard to accent classification by introducing a standard, open-source benchmark derived from Common Voice. We take the approach that implementing LAMs, without further modification to the architecture, yields promising results.

\begin{table}[t]
    \caption{Partition of train, dev and test sets. $^{\dagger}$also includes South Atlantic and Bermuda accented English. $^{\ddagger}$includes accents only from Italy.}
    \label{tab:cv-dataset}
    \centering
    \resizebox{\columnwidth}{!}{
    \begin{tabular}{l | p{4cm} | ccc}
        \toprule
        \rowcolor{Gray} \textbf{Language} & \multicolumn{1}{c}{\textbf{Accents}} & \multicolumn{3}{|c}{\textbf{\# Utterances [k]/dur [hrs]}} \\
        \cmidrule(lr){1-1} \cmidrule(lr){2-2} \cmidrule(lr){3-5}
        \textit{(Nb. accents)} &  & \textbf{Train} & \textbf{Dev} & \textbf{Test} \\
        \midrule
        English (16)$^{\dagger}$ & en-MY, en-SG, zh-HK, fil-PH, af-ZA, en-NZ, ga-IE, gd-GB, en-AU, en-CA, en-GB, en-IN, en-US, cy-CB, & 93.5/154 & 1.4/2.4  & 1.4/2.3  \\
        German (4) & de-IT, de-CH, de-AT, de-DE & 39.8/70  &  0.5/0.9 & 0.5/0.9  \\
        Spanish (6) & Mexico, Chile, Caribe, Rioplatense, Andino, Spain & 51.4/78  &  0.6/0.9 & 0.6/0.9  \\
        Italian (5)$^{\ddagger}$ & Romagna, it-Meridional, Veneto, Sicilia, Trentino & 2.6/3.7 & 0.37/0.6 & 0.37/0.6\\
        \bottomrule
        \end{tabular}
        }
\end{table}

\section{CommonAccent}
\label{sec:common_accent}

CommonAccent is a recipe that is derived from the Common Voice dataset~\cite{common_voice}. Common Voice is a massive multilingual corpus that contains annotations of speech recordings in over 100 languages (as of May 2023). The data is constantly updated with new recordings, therefore, the authors label each release with a number, e.g., \textit{Common Voice 11.0}. In addition to transcripts and language ID locales, some recordings contain information such as speaker's gender, accent, and age. Train, dev and test splits intended primarily for ASR are also provided.

This work employs the Common Voice dataset to perform accent classification on different languages. The CommonAccent recipe uses two versions of Common Voice. For English (EN), we use Common Voice 7.0 (CV7), while for German (DE), Spanish (ES), and Italian (IT) we use Common Voice 11.0 (CV11).\footnote{This recipe can be used with other datasets such as L2 Artic~\cite{zhao2018l2}.}

\noindent \textbf{Data selection process:} To prepare each dataset (e.g., EN) for analysis, all samples that are devoid of accent annotations are removed. Next, the train, dev, and test sets are combined into a single dataset since the original dev and test sets have an insufficient number of samples for some accents; in some cases, none at all. The number of samples per accent are then tallied to create new train, dev, and test sets, ensuring that no more than 100 samples per accent are included in the dev and test sets. If a given accent has fewer than 300 samples, the remaining samples are split using a 60/20/20 split for train, dev, and test sets, respectively. The data selection process is summarized in Figure~\ref{fig:dataset-selection}. CV7 is sourced directly from the CommonVoice website, while CV11 from the HuggingFace Hub~\cite{huggingface,huggingface_datasets}. The resulting train, dev, and test set proportions for each language in consideration are presented in Table~\ref{tab:cv-dataset}.\footnote{We open-source a data preparation script to parse the \mbox{CommonVoice} dataset from HuggingFace into a CSV file.}

\begin{figure}[t]
  \centering
  \includegraphics[width=0.95\columnwidth]{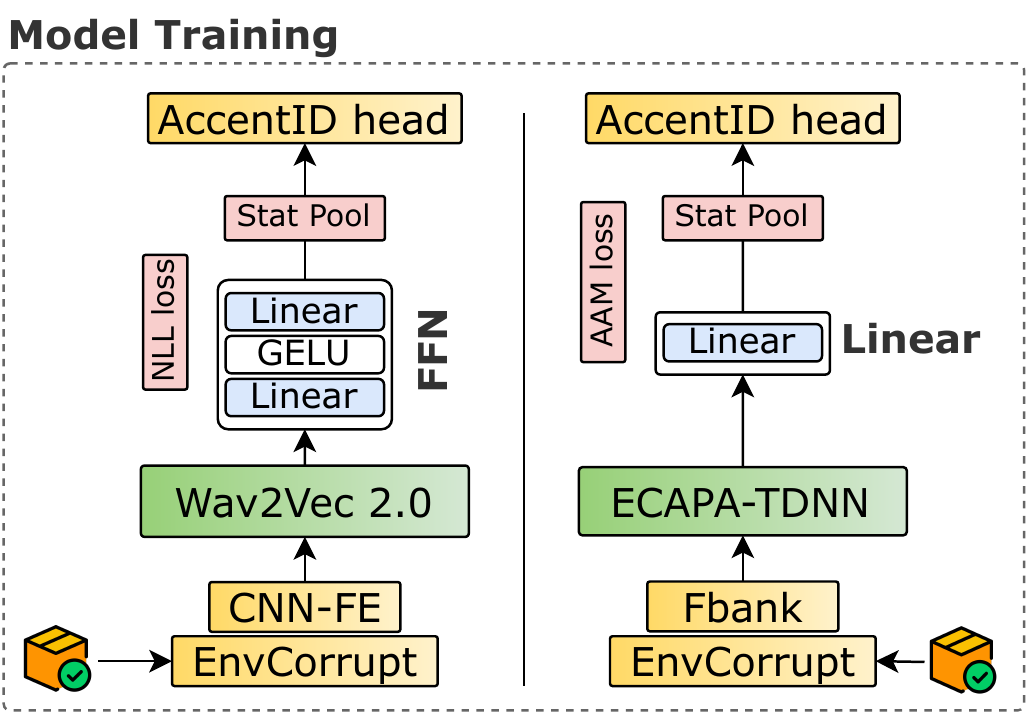}
  \caption{AccentID classification system. Model selection and training. We either fine-tune a pre-trained w2v2 (XLSR) or a ECAPA-TDNN model. The former uses NLL loss, while the later AAM loss. The w2v2 model is also interchangeable by other acoustic models. CNN-FE stands for Convolutional Neural Network Front-end. } 
  \label{fig:model-training}
\end{figure}

\section{Experimental setup}
\label{sec:experiments}

The proposed experimental setup is split in two parts:  1) fine-tuning w2v2-XLSR and ECAPA-TDNN models in order to obtain baseline results in English and 2) expanding the CommonAccent recipe for three additional languages using the w2v2-XLSR model~\cite{xlsr}. This serves the purpose of establishing that CommonAccent generalizes well regardless of different training scenarios with different pre-trained models and architectures. During experimentation, data augmentation by speed perturbation and additive noise from the OpenRIR database was applied~\cite{ko2017study_rir}.

\subsection{ECAPA-TDNN}

The ECAPA-TDNN model has shown state-of-the-art results in speaker verification tasks. It builds on the original x-vector architecture \cite{snyder2018x} through an increased focus on channel attention, propagation, and aggregation. The architecture includes an incorporation of Squeeze-Excitation blocks, multi-scale Res2Net features, extra skip connections, and channel dependent attentive statistics pooling as output~\cite{ecapa_tdnn}. 

We examine the implementation of this architecture in accent classification through two models: one trained with SpecAugmentation~\cite{Park2019} and speed perturbation and a baseline Accent Identification model without data augmentation. Both models are fine-tuned from the checkpoints on HuggingFace\footnote{The model we fine-tuned was based on Language Identification and was trained on the CommonLanguage dataset, see \url{https://github.com/speechbrain/speechbrain/tree/develop/recipes/CommonLanguage/lang_id} Checkpoint at HuggingFace: \url{https://huggingface.co/speechbrain/lang-id-commonlanguage_ecapa}} and we use additive angular margin loss~\cite{deng2019arcface}.



\subsection{Wav2Vec 2.0 \& XLSR}

The w2v2-XLSR model is designed to acquire cross-lingual speech representations for 53 languages, utilizing the raw waveform of speech to train on 56K hours of unlabeled data~\cite{xlsr}. Based on the w2v2 architecture \cite{wav2vec20}, it learns contextualized speech representations and multilingual quantized latent speech representations simultaneously. These shared representations facilitate cross-lingual knowledge transfer from high-resource languages to low-resource languages, thereby improving the performance of the latter. The w2v2-XLSR model has proven superior to prior approaches in both language identification and speaker identification tasks \cite{xlsr}.\footnote{Access to pre-trained w2v2-XLSR model checkpoint: \url{huggingface.co/facebook/wav2vec2-large-xlsr-53}} To merge the information across embeddings from the same utterance, we added a \texttt{StatPooling()} layer for both w2v2-XLSR models. Finally, we trained end-to-end with the NLL loss function. A global overview of the architecture can be found in the left panel of Figure~\ref{fig:model-training}.

\subsection{Training}

We perform two training strategies. The first one fine-tunes a pre-trained Wav2Vev 2.0/XLSR~\cite{wav2vec20,xlsr} model with NLL loss. The second one performs training with a ECAPA-TDNN network~\cite{ecapa_tdnn}. A workflow of the proposed architectures is on Figure~\ref{fig:model-training}. Adam~\cite{kingma2014adam} optimizer is used across all experiments, with a learning rate scheduler that anneals the learning rate ($\alpha=1e^{-3}$) after the end of each epoch ($\beta=0.95$). All models are trained for 30 epochs. Additionally, a dynamic batching strategy is used while training each model. Dynamic batching aims at reducing the amount of zero-padding in the input batches to the model. This in turn, reduces the overall training time. We use effective $max\_batch_{len}=600$ and $num_{buckets}=200$. At decoding time we use a fixed batch size of 16. All experiments run in a GeForce RTX 3090.

\begin{figure}[h!]
    \centering
    \begin{subfigure}{0.9\columnwidth}
        \includegraphics[width=1\columnwidth]{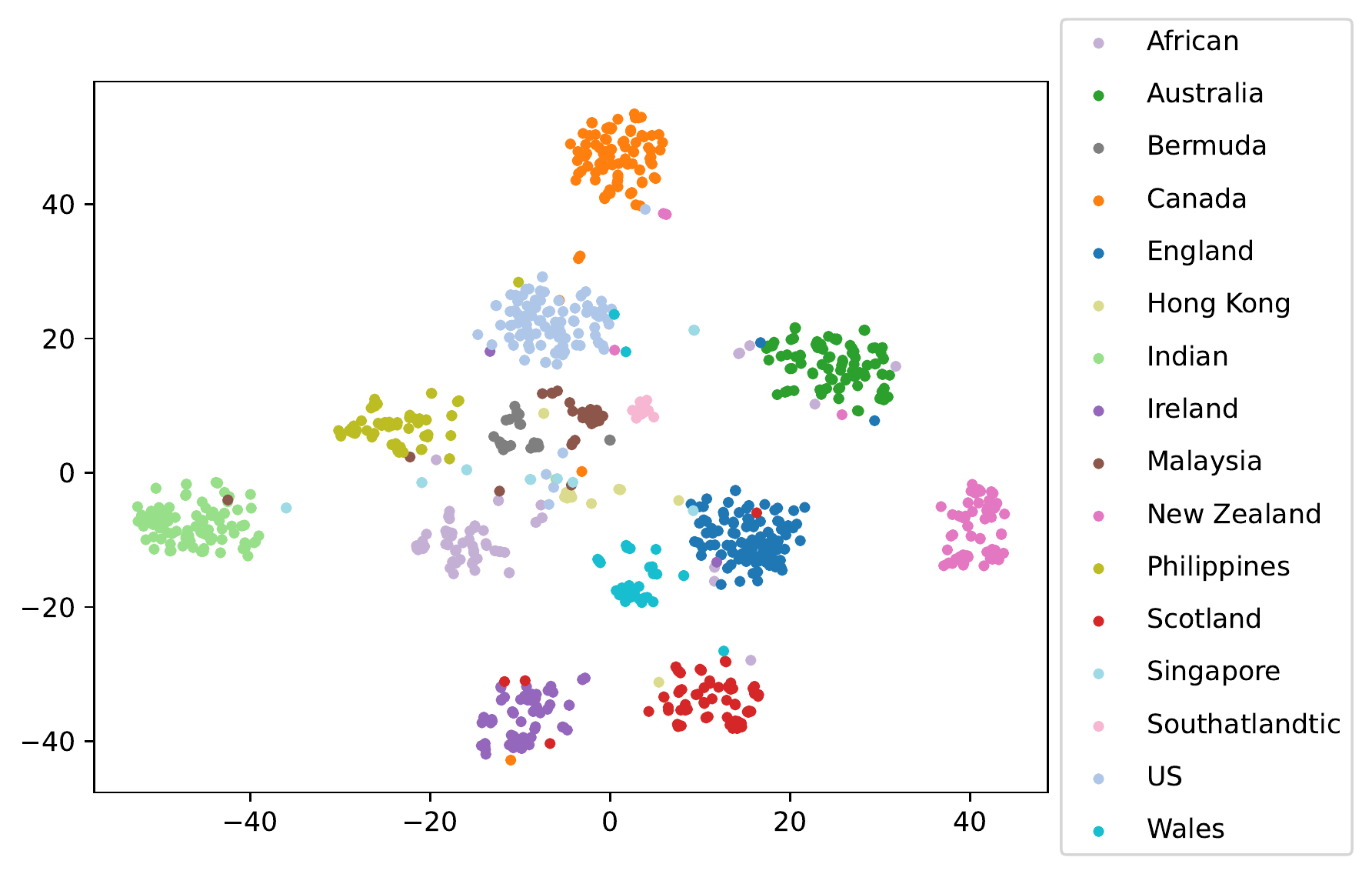}
        \caption{English. 16 accents. }
        \label{fig:sfig1}
    \end{subfigure}%
    \\
    \begin{subfigure}{1\columnwidth}
        \includegraphics[width=1\columnwidth]{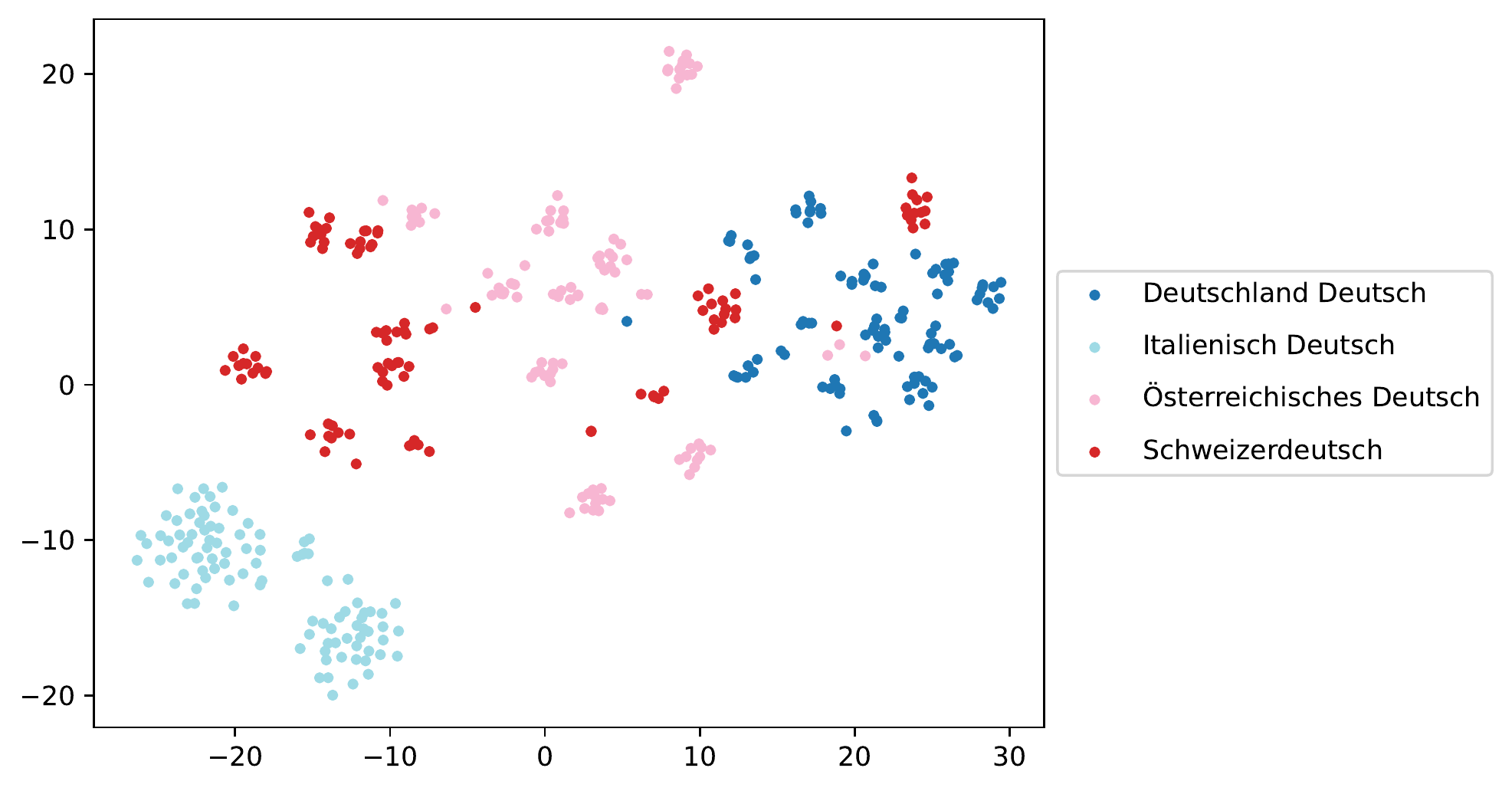}
        \caption{German. 4 accents. }
        \label{fig:sfig1}
    \end{subfigure}%
    \\
    \begin{subfigure}{1\columnwidth}
        \includegraphics[width=1\columnwidth]{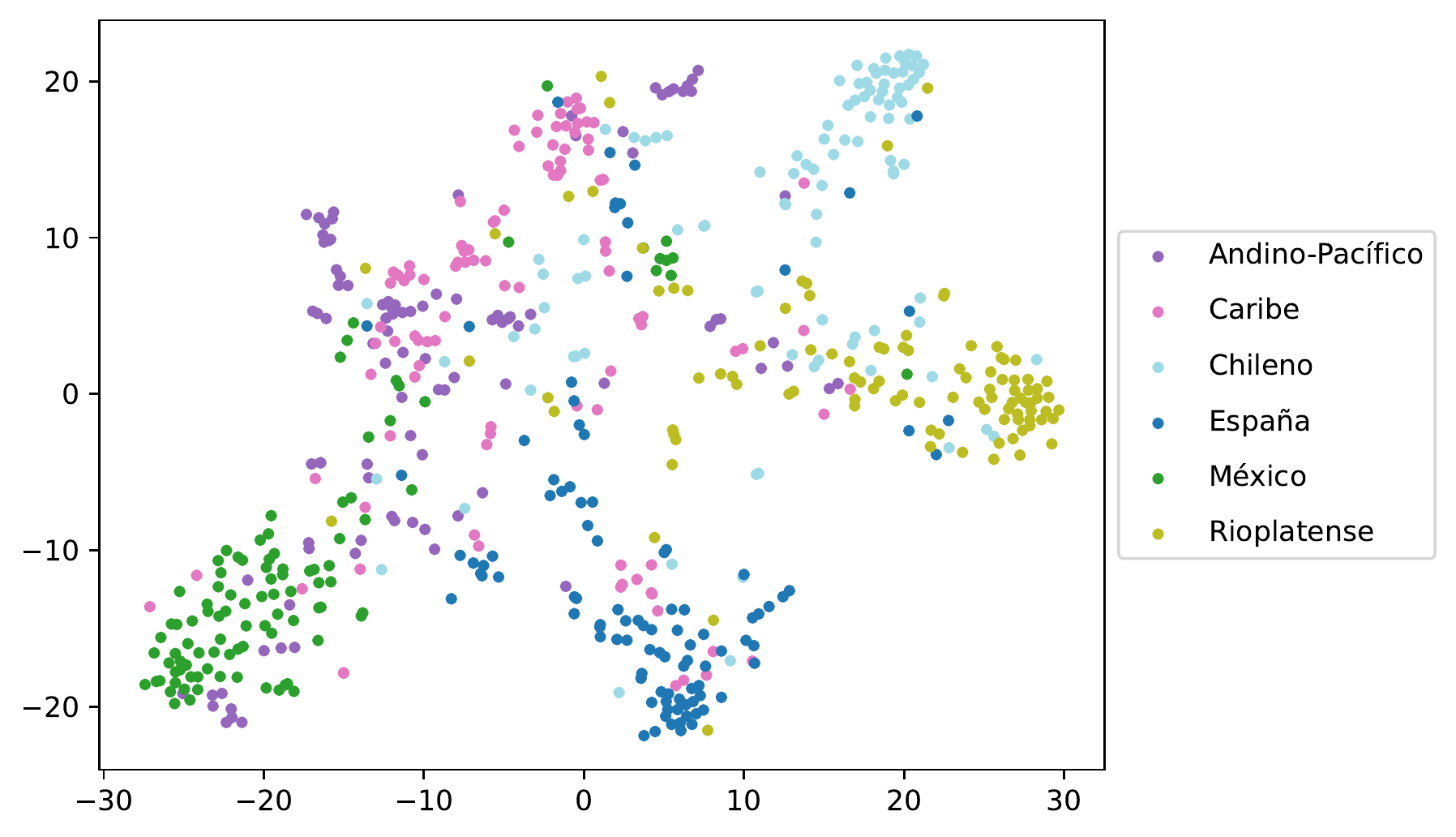}
        \caption{Spanish. 6 accents.}
        \label{fig:sfig1}
    \end{subfigure}%
    
\caption{Analysis of the internal categorization of Accent Classifiers by t-SNE plots. The plots list the internal categorization of (a) English, (b) German, and (c) Spanish. The embeddings are obtained from the last layer of the fine-tuned w2v2-XLSR model on each independent language (only test set). The embeddings have a dimension of 1024.}
\label{fig:t-sne-plots}
\end{figure}

\section{Results} 
\label{sec:results}

The results section is divided into three main parts. Firstly, we assess the accent classification performance of ECAPA-TDNN and w2v2-XLSR models on the English CommonAccent dataset (CV7), and conclude that the w2v2 pre-trained models are better suited for accent classification purposes. Secondly, we fine-tune these models for accent recognition in German, Spanish, and Italian. At the end, we conduct a clustering-based analysis of the embeddings derived from the fine-tuned w2v2-XLSR models.

\subsection{ECAPA-TDNN vs Wav2Vec 2.0}
\label{subsec:ecapa-vs-w2v2}

Table~\ref{tab:en-results} presents the accuracy scores for both ECAPA-TDNN and w2v2 models fine-tuned on the English CommonAccent dataset. The table clearly shows that the w2v2 models consistently outperform ECAPA-TDNN on the dev and test sets, with an accuracy improvement 79.0\% $\rightarrow$ 95.1\%. We also observe a significant improvement in accuracy for both models when applying SpecAugment technique. For instance, ECAPA-TDNN improves: 79.0\% $\rightarrow$ 89.7\% and w2v2-XLSR: 95.1\% $\rightarrow$ 97.1\%.

These results are not unexpected and demonstrate the effectiveness of data augmentation in addressing imbalanced data distributions, especially for low-resource accents within the same language. Additionally, the w2v2 model is less sensitive to data augmentation and consistently maintains a high level of accuracy. Perhaps this is due to the large-scale pretraining stage of XLSR. 

\begin{table}[t]
    \caption{Accuracy score on two type of models trained on the English CommonAccent dataset with and without data augmentation (speed and noise perturbation).}
    \label{tab:en-results}
    \centering
    \begin{tabular}{llll}
        \toprule
        \rowcolor{Gray} \textbf{Model} & \textbf{Aug.} & \textbf{Dev} & \textbf{Test} \\
        \midrule
        ECAPA-TDNN & \textcolor{red}{\XSolidBrush} & 78.5 &  79.0 \\
        $\hookrightarrow$ & \textcolor{teal}{\checkmark}  & 91.5 & 89.7  \\
        \midrule \midrule
        w2v2-XLSR & \textcolor{red}{\XSolidBrush} & 95.2 &  95.1 \\
        $\hookrightarrow$ & \textcolor{teal}{\checkmark}  &  96.5 & 97.1 \\
         \bottomrule
    \end{tabular}
\end{table}

\subsection{Baselines In Other Languages}

To further analyze these results, a series of experiments were conducted by fine-tuning the w2v2-XLSR models on the German, Spanish, and Italian CommonAccent datasets. Based on the superior results of the w2v2-XLSR model from \S~\ref{subsec:ecapa-vs-w2v2} we only continue studies with this model on the other 3 languages and summarize the results in Table~\ref{tab:es-de-it-results}. Although the w2v2-XLSR model yielded above 95\% accuracy for English, substantial degradation for other languages can be seen. Overfitting was observed in the Italian model (76.1\% dev $\rightarrow$ 99\% test) and underfitting in the Spanish model (below 69\% accuracy on dev and test sets). One possible explanation is the high degree of phonological similarity between the accents, which makes the distinction more subtle. For example, five out of the six Spanish accents are from closely located regions (Latin America), which could be a contributing factor to the model's low performance. However, the German model performed relatively well, achieving an accuracy of over 75\%. We also used t-SNE plots (see Figure~\ref{fig:t-sne-plots}) to visualize the clustering of the accents, which clearly show that the models are learning meaningful representations of the accents within the same language.

\begin{table}[t]
    \caption{Accuracy score in German, Spanish, and Italian CommonAccent datasets. Results are with a w2v2-XLSR model fine-tuned on each language independently. $^{\dagger}$includes speed perturbation and noise perturbation during fine-tuning.}
    \label{tab:es-de-it-results}
    \centering
    \begin{tabular}{l c c c c}
        \toprule
        \rowcolor{Gray} \textbf{Locale} & \textbf{Aug.$^{\dagger}$} & \textbf{Dev} & \textbf{Test} & \textbf{Test loss} \\
        \midrule
        German & \textcolor{teal}{\checkmark} & 66.2 & 75.5 & 0.937 \\
        Spanish & \textcolor{teal}{\checkmark} & 64.2 & 68.5 & 1.22  \\
        Italian & \textcolor{teal}{\checkmark} &76.1  & 99.0 & 0.392 \\
         \bottomrule
    \end{tabular}
\end{table}

\subsection{Clustering}

As shown in Figure 3, the embeddings for each language in the w2v2-XLSR architecture show a level of clustering based on phonological similarity and geographical proximity. For example, the English plot displays England, Wales, and Scotland close to each other. US and Canada are also next to each other as is Australia and New Zealand. While the t-SNE plot shows that the model has recognized similarities in accents, the relational differences between them is not so clear. For example, the Australian accent is equidistantly placed between US and New Zealand, although there are considerable difference between the IPA phonemes for US English, while Australia and New Zealand are the same.\footnote{As given in the IPA charts for Amazon Polly: \url{https://docs.aws.amazon.com/polly/latest/dg/ref-phoneme-tables-shell.html}}

\section{Conclusion}
\label{sec:conclusion}

In this work, we showed that pre-trained acoustic models like XLSR can be adapted for accent classification systems, particularly for English (\S~\ref{sec:results}). We also introduced CommonAccent (\S~\ref{sec:common_accent}), a benchmark dataset for accent classification tasks in four languages. We hope it becomes the default benchmark, like VoxCeleb for speaker recognition, as there is currently a lack of standardized benchmarks within the domain targeted in this paper. Finally, we open-sourced our models in English, Spanish, German, and Italian, along with the accent classification recipe in SpeechBrain that can be easily adapted to other datasets and languages.

\bibliographystyle{IEEEtran}
\bibliography{biblio}

\end{document}